\documentclass[10pt,letterpaper]{article}

\usepackage{ccn}
\usepackage{float}
\usepackage{dblfloatfix}
\usepackage{pslatex}
\usepackage{apacite}
\usepackage{soul}
\usepackage[dvipsnames]{xcolor}
\usepackage{graphicx}% http://ctan.org/pkg/graphicx

\title{Diagnosing Catastrophe: Large Parts of Accuracy Loss in \\Continual Learning Can Be Accounted for by Readout Misalignment}

\author{{\large \bf Daniel Anthes \qquad Sushrut Thorat \qquad Peter König \qquad Tim C. Kietzmann} \\
danthes@uos.de, sushrut.thorat94@gmail.com, pkoenig@uos.de, tkietzma@uos.de\\
  Institute of Cognitive Science, Osnabrück University}

\begin{document}

\maketitle

% \clearpage
% \newpage

\begin{figure*}[!t]
  \centering
  \includegraphics[width=\textwidth]{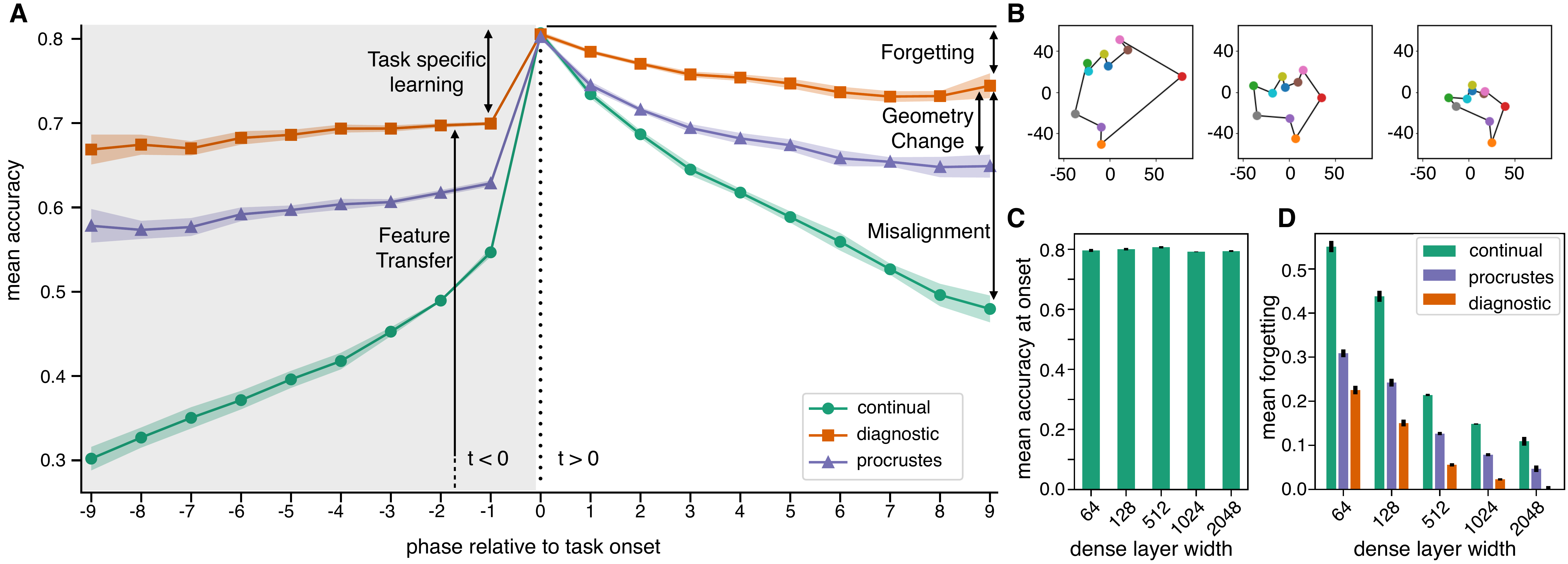}
  \caption{\textbf{A:} Classification accuracy averaged over the ten tasks sampled from CIFAR100. Prior to averaging, task performance trajectories are temporally aligned to task onset such that the x-axis reflects performance after t additional tasks have been learned. The shaded area around each line indicates the standard error computed over five repetitions of the procedure. \textbf{B:} Mean class representations for a task split. 
  Task mean representations from all phases are projected to a shared two dimensional space using multidimensional scaling~\cite{torgerson1952multidimensional}. 
  Shown are representation vectors directly after learning the task, after learning 5, and after learning 9 additional tasks (left to right). \textbf{C:} Mean performance over all tasks at task onset ($t=0$). Network size does not have an effect on how well tasks are learned initially. \textbf{D:} Performance loss measured as the difference between performance at $t=0$ and the mean over performances measured at all $t > 0$. Standard error for additional networks is computed over three simulations.} 
\label{fig1}
\end{figure*}

\section{Abstract}
{
\bf Unlike primates, training artificial neural networks (ANNs) on changing data distributions leads to a rapid decrease in performance on old tasks. This phenomenon is commonly referred to as catastrophic forgetting. In this paper, we investigate the representational changes that underlie this performance decrease and identify three distinct processes that together account for the phenomenon. The largest component is a misalignment between hidden representations and readout layers. Misalignment occurs due to learning on additional tasks and causes internal representations to shift. Representational geometry is partially conserved under this misalignment and only a small part of the information is irrecoverably lost. All types of representational changes scale with the dimensionality of hidden representations. These insights have implications for deep learning applications that need to be continuously updated, but may also aid aligning ANN models to the rather robust biological vision.
}
\begin{quote}
\small
\textbf{Keywords:} 
continual learning; catastrophic forgetting; artificial neural networks; representations; lifelong learning
\end{quote}

\section{Introduction}

Our world is inherently sequential. Adapted to this, humans are successful in continuously learning new skills over their lifetime. However, most state-of-the-art training procedures for artificial neural networks (ANNs) rely on data being independent and identically distributed. In settings where the data distribution changes, networks have been reported to rapidly forget previous knowledge~\cite{parisi2019continual, hadsell2020embracing}. This phenomenon is commonly termed \textit{catastrophic forgetting}~\cite{french1999catastrophic,mccloskey1989catastrophic}.

A number of factors influence the degree to which performance decreases in sequential learning scenarios: the dimensionality of representations~\cite{mirzadeh2022wide}, pretraining~\cite{ramasesh2022effect}, objective function~\cite{li2022energy, davari2022probing} and task similarity~\cite{ramasesh2020anatomy}. However, the changes to the task-relevant representations during continual learning remain to be fully characterized (see \citeA{davari2022probing} for first steps). In this work, we characterize changes in representational geometry and their contribution to the observed decrease in performance. We find that rather than forgetting, much of the degraded performance can be explained by a misalignment of representations and the readouts of the network.

\section{Analysis}

Our model system is a standard four layer convolutional network
%\footnote{The network training and analysis scripts can be found at \url{https://github.com/KietzmannLab/ReadoutMisalignment}}. 
The training procedure, task and network architecture are identical to \citeA{zenke2017continual}. We study catastrophic forgetting in the task-incremental scenario~\cite{van2019three}, initializing a new classification head every time a novel task is encountered. 
After pretraining the network on CIFAR10~\cite{krizhevsky2009learning}, we sequentially train on ten equal task splits from CIFAR-100. We repeat this procedure $5$ times controlling for the effects of task similarity by randomly assigning each class to a task~\cite{ramasesh2020anatomy}.
% Previous work has shown that catastrophic forgetting concentrates in later layers \cite{ramasesh2020anatomy}. Therefore we focus our investigations on changes in the dense pre-readout layer.

%There are two distinct ways in which representations (pre-readout activations) can change during learning: one, the discriminability between the classes can change. Two, the representational geometry can change. This change can preserve or distort the geometry of the class representations. From the perspective of the readout, these changes can misalign the representations with the corresponding readout directions.
%TCK This paragraph is not clear to me. You start by saying the two options are distinct. However, the first option, class discriminability, will also change the geometry. So what is the difference between the two exactly?  Is one geometry preserving and one not? Your current second option does imply both again. In short, the above paragraph does not make sense right now, so maybe try and express more clearly what you mean.

We characterise the information present throughout learning by training diagnostic readouts for all tasks after every phase of training. A drop in performance, despite adjusted readout, constitutes a loss of task relevant information. This scenario constitutes true \textit{forgetting}.
Contrary to this, performance loss attributed to \textit{misalignment} is computed by the difference in performance between the original readout ($t=0$) and the newly trained diagnostic readouts at every phase of training.
Third, to estimate the extent to which misalignment is due to rotation, translation, and uniform scaling of an otherwise static geometry, we align representations for each task after each training phase to the representations immediately after learning the task ($t=0$) with a geometry-preserving Procrustes transformation~\cite{gower1975generalized}. 
%Measuring the performance of the aligned representations with the original task readout layer and comparing it with the performance at $t=0$ is indicative of how much the representational geometry changed.

Finally, as increasing layer width has been shown to alleviate catastrophic forgetting~\cite{mirzadeh2022wide}, we vary the width of the final hidden layer to investigate how the different components of representational change are modulated by network capacity.

\section{Results}

% here we can be more elaborate
As expected, we observe effects of catastrophic forgetting, i.e. a rapid drop in performance of the original readouts as the network is trained on additional tasks (Fig.~\ref{fig1}A, `continual' at T$>$0). Notably, however, performance of diagnostic readouts decreases much less, indicating that the discriminability of the old classes is indeed preserved, i.e. there is little ``actual'' forgetting. The primary cause of decreased performance is readout misalignment, the extent of which is shown by the large difference between `continual' performance and performance measured at the diagnostic readouts (Fig.~\ref{fig1}A, `diagnostic'), in line with similar previous analyses ~\cite{davari2022probing}.

Does misalignment preserve the original representational geometry? If so, we'd expect that Procrustes alignment should yield performance as good as the linear diagnostic readouts. We observe that aligning representations accounts for approximately half of the performance difference between continual and diagnostic readouts (Fig.~\ref{fig1} A, `procrustes'). Therefore, misalignment can be characterized as a combination of geometry preserving and deforming changes of representations.

An open question that remains from our and previous work is whether the comparably good performance of the diagnostic readout is explained by transfer learning based on features learned for earlier tasks in the sequence. Indeed, we observe that the features learned for previously encountered tasks transfer to unseen tasks (`Feature Transfer' in Fig.~\ref{fig1}). Yet, transfer cannot fully explain the performance observed with diagnostic readouts, as a clear discontinuity in the diagnostic readout performance trajectory from before to after training a new task ($t=0$) can be seen. This suggests that newly learned features better support the new task. This additional information stays preserved in the network over learning of multiple additional tasks, as evidenced by the fact that diagnostic readout performance stays above the performance measured at $t=-1$ for the subsequent phases ($t>0$).
% Learning additional tasks prior to learning a particular task under consideration does not increase diagnostic readout performance (the orange line in Fig.\ref{fig1} is mostly flat for $t<0$) indicating that there is no accumulation of `transferable features' as additional tasks are learned.

Finally, characterizing the influence of network size in continual learning with our new analysis techniques, we find that varying the width of the final hidden layer attenuates all three measures of representational change. Yet, we still observe small amounts of changes to the representational geometry and misalignment with the readouts of the respective networks (Fig.~\ref{fig1} C \& D).
%TCK: Not even an eagle has eyesight good enough to decipher your tiny font in C and D. You need to make both panels substantially larger. I have shortened the discussion a little bit, which means that you can make the figure a but taller. I have sent a suggestion for a rearrangement of the figure in Basecamp. Idea: Make panel A a bit smaller, and make B horizontal, C and D below.

\section{Discussion}

In characterizing representational changes in a neural network during continual learning, we observed that misalignment of the pre-readout representations with the task readouts explains large parts of performance degradation that is commonly referred to as `catastrophic forgetting'. Interestingly, only a small amount of performance cannot be linearly read out and is irrecoverably `forgotten'.

Many algorithms addressing catastrophic forgetting rely on restricting learning at synapses that encode information for previous tasks \cite{zenke2017continual, kirkpatrick2017overcoming} or regularize learning of representations for new tasks \cite{li2017learning} in order to not lose information relevant for the previous tasks. We argue that information in hidden layers is largely preserved, even without restricting learning trajectories or placing constraints on representations the network is allowed to learn. This is especially prominent in larger networks. We hypothesize that catastrophic forgetting may instead be efficiently addressed by solving the problem of readout misalignment without influencing the learning of new tasks (See also: \cite{lesort2021continual}). Indeed, there may be benefits to not restricting learning of representations more than necessary, as restrictions to the learning dynamics of the network may lead to decreased plasticity or sub-optimal solutions over long sequences of tasks.
% I would write that for ANNs to be models of the brain, the issue clearly needs to be addressed. If the brain does change with new tasks, too, then there is a strategy to cope with it. Misalignment may be one way to track change and deal with the problem.

Lastly, the primate visual system is successfully able to learn new tasks without exhibiting forgetting of old tasks. If we are to use ANNs as models of biological vision, then the discrepancies in the learning dynamics of the two systems remain to be addressed. Future work will test the currently described analysis framework for characterising representational changes in continual learning on biological data to further understand where, how, and when the visual system copes with the newly arriving information.

% Researching how misalignment is dealt with in these systems may be a fruitful source of inspiration for solving readout misalignment in continual learning.
% \vfill
\section{Acknowledgments}
The project was financed by the Deutsche Forschungsgemeinschaft (DFG, research training group ``Computational Cognition'', GRK2340), as well as the European Research Council (ERC, TIME, Project 101039524).
Compute resources used for this project are funded by the Deutsche Forschungsgemeinschaft (DFG, German Research Foundation), project number 456666331.
% \newpage
\bibliographystyle{apacite}

\setlength{\bibleftmargin}{.125in}
\setlength{\bibindent}{-\bibleftmargin}

\bibliography{ccn_style}

\end{document}